\documentclass[10pt,twocolumn]{article}

\usepackage{wacv}
\usepackage{times}
\usepackage{epsfig}
\usepackage{graphicx}
\usepackage{amsmath}
\usepackage{amssymb}
\usepackage{balance}

\usepackage[pagebackref=true,breaklinks=true,colorlinks,bookmarks=false]{hyperref}

\wacvfinalcopy 
\def\wacvPaperID{218} 

\ifwacvfinal\fi
\begin{document}

\title{Fair Comparison: Quantifying Variance in Results \\for Fine-grained Visual Categorization}

\author{Matthew Gwilliam\textsuperscript{1,2}\quad Adam Teuscher\textsuperscript{1} \quad Connor Anderson\textsuperscript{1} \quad Ryan Farrell\textsuperscript{1} \\
\textsuperscript{1}Brigham Young University, \textsuperscript{2}University of Maryland\\
{\tt\small \{mattgwilliamjr,adam.m.teuscher\}@gmail.com} \\
\tt\small connnor.anderson@byu.edu \quad farrell@cs.byu.edu
}

\maketitle

\begin{abstract}
For the task of image classification, researchers work arduously to develop the next state-of-the-art (SOTA) model, each bench-marking their own performance against that of their predecessors and of their peers. Unfortunately, the metric used most frequently to describe a model's performance, average categorization accuracy, is often used in isolation. As the number of classes increases, such as in fine-grained visual categorization (FGVC), the amount of information conveyed by average accuracy alone dwindles. While its most glaring weakness is its failure to describe the model's performance on a class-by-class basis, average accuracy also fails to describe how performance may vary from one trained model of the same architecture, on the same dataset, to another (both averaged across all categories and at the per-class level). We first demonstrate the magnitude of these variations across models and across class distributions based on attributes of the data, comparing results on different visual domains and different per-class image distributions, including long-tailed distributions and few-shot subsets. We then analyze the impact various FGVC methods have on overall and per-class variance. From this analysis, we both highlight the importance of reporting and comparing methods based on information beyond overall accuracy, as well as point out techniques that mitigate variance in FGVC results.
\end{abstract}

%
%
\section{Introduction} \label{introduction}

\begin{figure}[ht!]
\centering
\includegraphics[width=\linewidth]{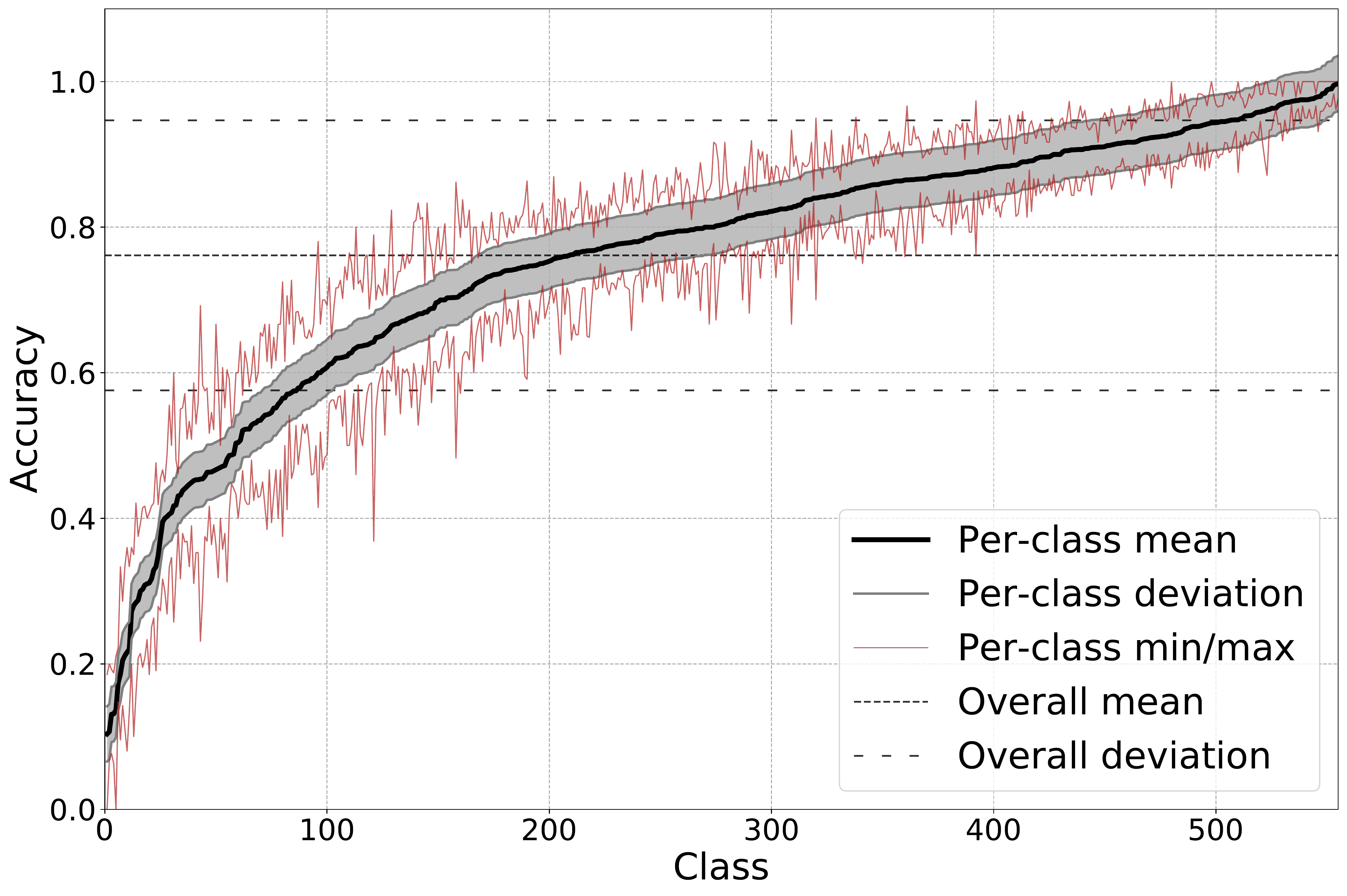}
\caption{Results from 10 trained networks in per-class accuracy plot, sorted according to the average per-class accuracies. Deviation is standard deviation, and the per-class deviation was averaged across all the classes.}
\label{intro_fig}
\end{figure}

Convolutional neural networks and their variants have taken over the image classification task, both at the general level~\cite{RussakovskyDSKSMHKKBBF_IJCV2015} and for fine-grained domains~\cite{BransonVBP_BMVC2014,ZhangDGD_ECCV2014}. This has advanced the state-of-the-art in fine-grained recognition~\cite{Chen_2019_CVPR,DubeyGRN_NeurIPS2018,Ge_2019_CVPR,LinRM_ICCV2015,Luo_2019_ICCV,WangMD_CVPR2018,ZhengFML_ICCV2017} to previously inconceivable levels, but with great research comes a need for great reporting, and current standards are insufficient. If a given model yielded the same results every time it was trained, and these results were the same for every class, then overall average accuracy would be not only sufficient, but also comprehensive, and additional metrics would be mostly trivial. Unfortunately, this is not the case, and as FGVC transitions from theory to real-world application, it is critical that both researchers and practitioners have enough information at their disposal to fairly compare different models.

One of the primary limitations in comparing FGVC methods is that models with the exact same dataset, training procedure, and architecture (identically-trained models) have non-trivial differences (variance) in performance. Variance is an important concern about a model's performance that is distinct from accuracy -- the level of variance can vary independent of overall accuracy. We investigate a few main types, two of which are represented in Figure \ref{intro_fig}. The first, which we measure as \textbf{overall deviation} in Section~\ref{var_by_data} and Section~\ref{var_by_method}, is the wide distribution of per-class accuracies: for some classes, the trained model is able to classify nearly all the images correctly, while for a few classes it accurately classifies less than 20\%. The second type of variance, which we measure as \textbf{per-class deviation} in Section~\ref{var_by_data} and Section~\ref{var_by_method} is the distribution of accuracies for a given class between trained instances of a model. The third type, which is not represented in the figure, is variance in overall accuracy. Ironically, the variance between results from different trained instances is due primarily to what they have in common: \textbf{architecture}, training \textbf{procedure}, and \textbf{dataset}. 

The \textbf{architecture} is what makes variance possible. 
Deep neural networks have a series of layers, with millions of parameters in total, typically initialized at random. 
Since there are so many learned weights, the possible solution space is enormous, and due to the way these weights are learned and the way they work together, even a reduced space consisting only of ``good'' solutions would still be incredibly large. 
With the introduction of any randomness in the training pipeline (explained in the following paragraph), the learned weights will differ even when not randomly initialized (such as with pretrained networks) and trained models become like snowflakes, in that it is highly unlikely any two would ever have the exact same weights. 
Two trained models with non-identical weights would logically make predictions that, while largely similar, have some differences (variance).

In terms of training \textbf{procedure}, while every new state-of-the-art (SOTA) approach to FGVC is different, many share some similarities in their pipelines. 
The order of training images presented during training is often shuffled, with transformations applied that randomly flip, crop, and modify the color of the images in each batch.
While the addition of randomness to the pipeline through these methods helps prevent overfitting, it allows for variance between instances of a model that started off with identical weights.

\begin{figure}[t!]
\centering
\includegraphics[width=\linewidth]{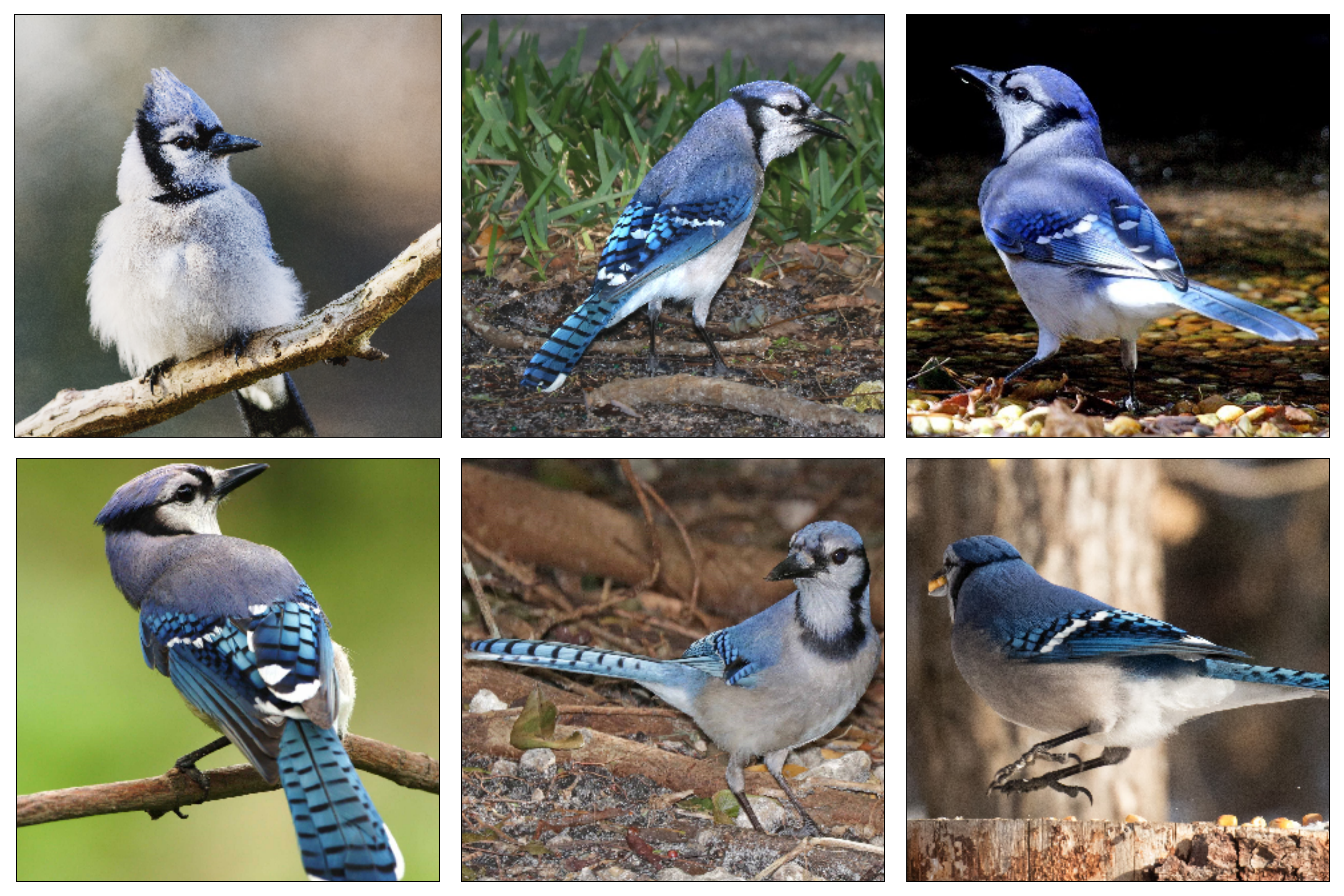}
\caption{\textbf{Blue Jays}, from Flickr. Differences in pose make it difficult to discern (without prior domain knowledge) whether these are all images of the same species of bird.}
\label{blue_jays}
\end{figure}

\begin{figure}[t!]
\centering
\includegraphics[width=\linewidth]{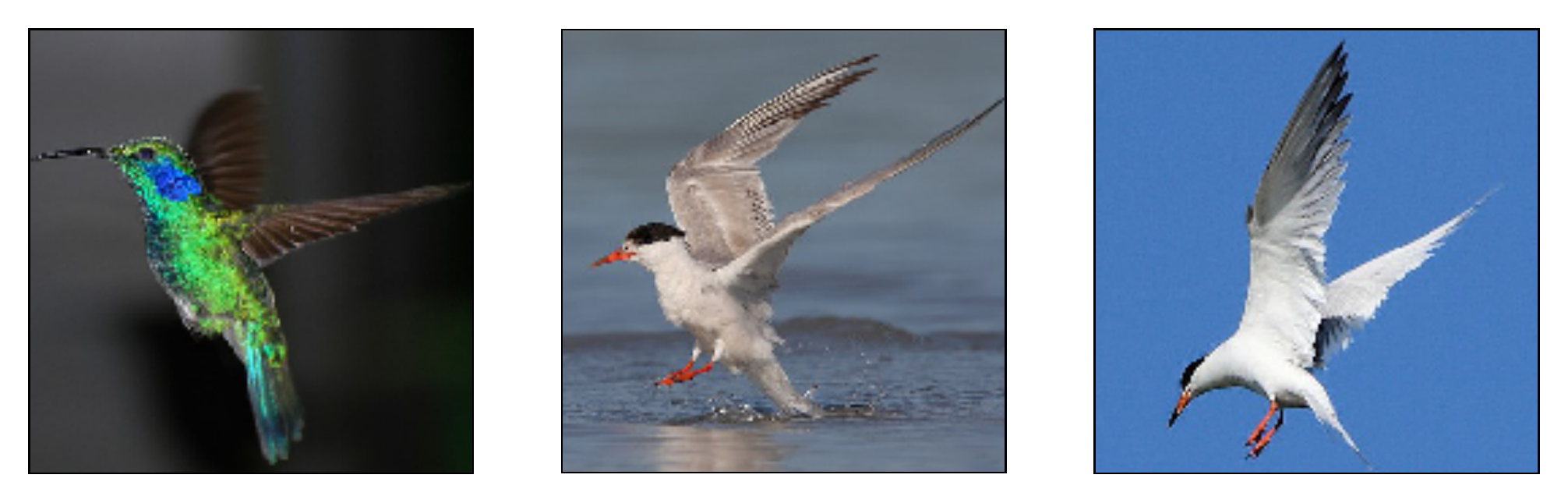}
\caption{From left to right: \textbf{green violetear}, \textbf{common tern} and \textbf{elegant tern}. In CUB, the violetear is visually distinct, while the common and elegant terns are easily confused.}
\label{three_birds}
\end{figure}

The \textbf{dataset} is a major factor in determining the level of variance observed in the evaluation results, as we prove in Section~\ref{var_by_data}. 
It is important to remember that the training images in a dataset are a random sample taken from the space spanned by the selected visual domain, and a model's performance can be highly dependent on how representative this random sample is of the full visual domain. 
If there are only a few images for a given class in the training set, it is unlikely that the spectrum of possible appearances for that class is adequately represented.
Figure~\ref{blue_jays} illustrates how this can be difficult -- with only 5 images, it would be impossible to represent each of the 6 poses shown, and even with more than 20 images it would logically be challenging to achieve adequate, well-proportioned representation of the Blue Jay.
With a limited training set, performance on the test data will likely be both (i) worse for the underrepresented classes and (ii) less consistent overall.
Additionally, some classes in a dataset are more challenging to classify than others.
For example, with the 2011 Caltech-USCD Birds (CUB) dataset~\cite{WahCUB_200_2011}, many models report very high accuracy for highly-distinctive birds like the green violetear but lower accuracy for birds that are hard to differentiate, such as the common and elegant terns, shown in Figure \ref{three_birds}.

Variance can be seriously problematic in a few different ways. 
The wideness of the distribution of per-class accuracy scores presents challenges for a multitude of practical use cases.
Consider the distribution in Figure~\ref{intro_fig} as an example, and note while some per-class accuracies are near 100\%, other are close to 0\%.
An x-ray classifier with such a distribution would often fail to classify certain types of breaks and fractures. 
Such a bird classifier may never be sufficiently accurate for the endangered species in the domain.
Ensemble methods can turn some of the per-class variance into a strength, but due to resource-consumption, such an approach may be impractical for many applications, making variance in performance between trained instances of the same model an important consideration.
For some use cases, when two models have similar average accuracies, it may be better to use the model with less variance between trained instances.
However, since data that describes the variance is often not reported, it is difficult for FGVC practitioners to make such decisions in an informed manner.

Variance also makes fair reporting and comparison of results quite challenging. Consider two fine-grained researchers, A and B, who develop two different methods for the classification task. Researcher A, in order to make sure their results are reproducible, writes a paper where they report an overall accuracy that comes from the average of 10 trials. Researcher B runs 10 trials as well, and unfortunately, the mean accuracy of their results is lower than that of Researcher A. However, 2 of their trials achieve higher accuracies than Researcher A's reported accuracy. Perhaps unaware that Resarcher A was reporting an average over several trials, Researcher B reports the maximum from the 10 trials as their overall accuracy. Other readers, not knowing which methodology was used to get the overall accuracy figure (since it is often not reported or hard to find), assume that in terms of overall accuracy the method of Researcher B outperforms the method of Researcher A, when in reality, Researcher A's method is better on average. This and other potential issues can be observed in the experiments we run using real methods and real data in Section~\ref{var_by_data}.

Therefore, variance presents a major stumbling block for the fair reporting, comparison, and practical application of different FGVC models.
In this paper, we draw attention to the variations in performance based on many different factors. 
We break down and reveal trends describing how variations in overall accuracy, per-class accuracy, and range of per-class accuracies can be influenced by visual domain (dataset), shape of a long-tailed distribution, optimizer, model architecture, loss function, and number of training images per class.
We both demonstrate the prevalence of variance and uncover insights on what drives it.

%
%
\section{Background and Related Work} \label{related_work}

One recent and striking example of the variance in model performance caused by random initialization is the lottery ticket hypothesis~\cite{frankle2018lottery}. The hypothesis states that for a dense, randomly initialized network there exists a sub-network with fewer parameters that can attain the same or better accuracy in the same or less amount of time when trained independently on the same task. These sub-networks are dubbed ``winning tickets", because it is the unique combination of initial weights and sub-network structure that allow them to learn effectively. The authors observed that when the sub-networks were re-initialized with different random parameters, they failed to achieve the same level of performance. This precise dependence on the random initialization is evidence of the large variation that can occur when training neural models.

As discussed throughout Section \ref{introduction}, variance is an important subject in FGVC, where classification algorithms seek to identify the precise category/class within a particular domain (e.g. genus/species of a bird or make/model/year of a car). FGVC datasets such as those mentioned earlier frequently have limited numbers of images per category. Each image therefore contributes proportionally more toward overall accuracies (and variations therein). Near state-of-the-art approaches for FGVC include methods that use pooling~\cite{CuiZWLLB_CVPR2017,LinRM_PAMI2017}, attention mechanisms~\cite{FuZM_CVPR2017,ZhengFML_ICCV2017} or regularization~\cite{DubeyGRN_NeurIPS2018}.

Accuracy is the standard metric for comparing results across fine-grained classification models. 
However, not all deep learning disciplines are able to use such a simple metric. Machine translation (MT), a subfield of natural language processing, is one such discipline.
Similar to FGVC, MT has seen exciting advances in the past decade with the development of Neural MT methods~\cite{bahdanau2014neural,sutskever2014sequence,vaswani2017attention}.
Unlike FGVC, however, translation is an inherently subjective task, where there is no single right answer, and preferences vary between different human evaluators~\cite{DBLP:journals/corr/abs-1808-10432}.

As a result, MT practitioners have had to be resourceful in crafting their evaluation metrics.
The most popular automatic metric is BLEU score~\cite{Papineni:2002:BMA:1073083.1073135}, which can be calculated using frameworks such as SacreBLEU~\cite{post-2018-call}.
Unfortunately, simple comparison of BLEU scores is inferior to human evaluation.
To try to bridge the gap between human and machine evaluation of machine translation, recent work has focused on producing additional metrics that measure things like linguistic richness~\cite{vanmassenhove-etal-2019-lost} or level of formality~\cite{niu-etal-2017-study} in translated text. 
We take a similarly creative approach in our proposed evaluation of FGVC systems. 

The most significant resource that we are aware of that addresses the fair reporting and comparison gap in FGVC is paperswithcode.com~\cite{papers_with_code}, which facilitates the comparison of common metrics like top 1 and top 5 accuracy, with links to both the accompanying paper and code. Our work also bears some similarity to general machine learning benchmark suites, such as the UCI Machine Learning Repository~\cite{Dua:2019,macia2014towards} and the Penn machine learning benchmark~\cite{article}.
Instead of focusing merely on accuracy, we propose additional metrics and ways to evaluate the performance of classifiers.

%
%
\section{Variance by Data} \label{var_by_data}
\begin{figure*}[ht!]
\centering
\includegraphics[width=\linewidth]{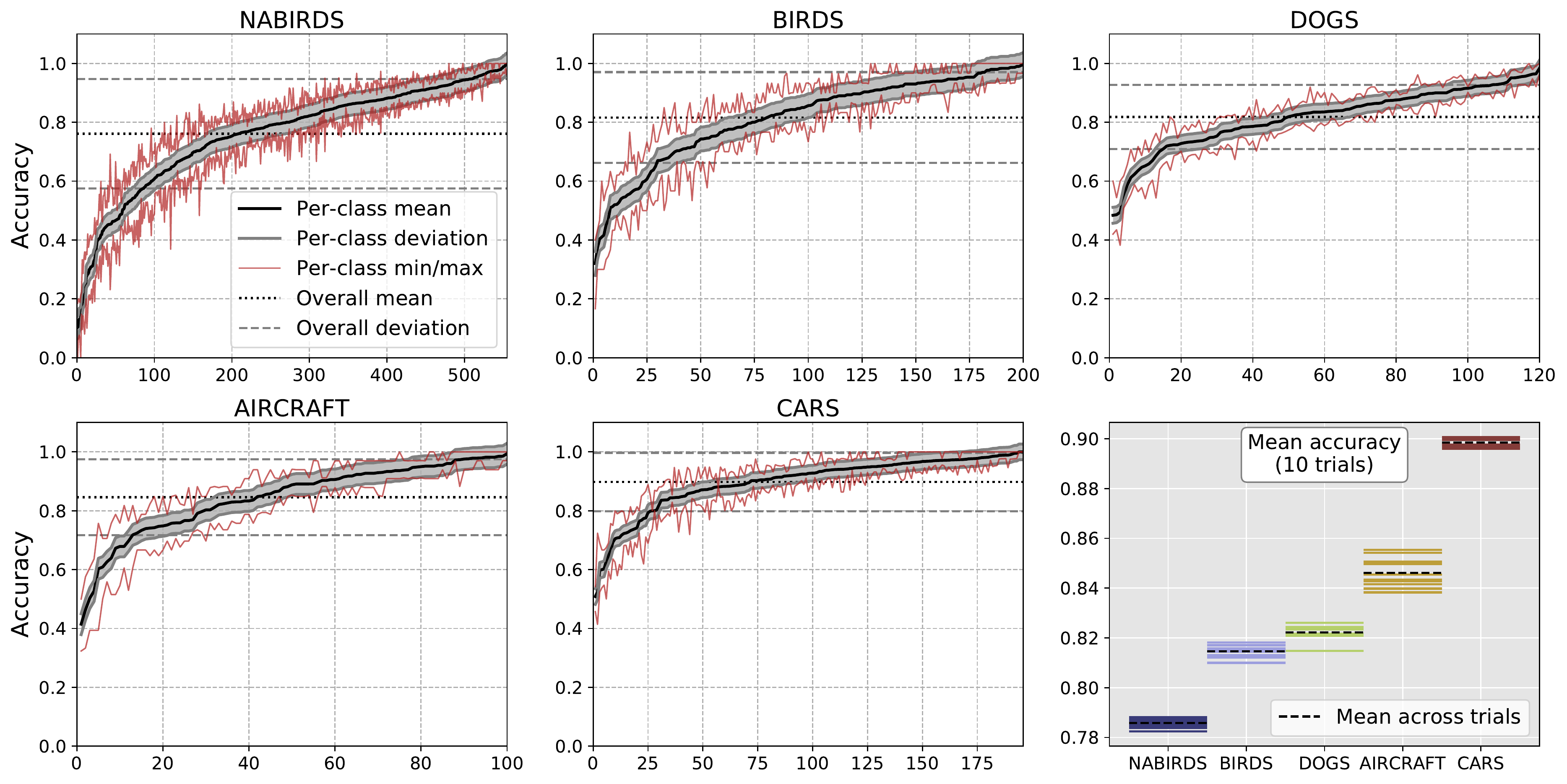}
\caption{Per-class accuracy for each dataset, aggregated across 10 training runs. The standard deviation (gray band) and minimum/maximum performance (red-colored curves) for each class are shown along with the mean (black curve). The overall mean and standard deviation across classes are also shown (dotted lines). \textbf{Bottom right}: Overall accuracy for each trial.}
\label{dataset_accs}
\end{figure*}

The experiments in this section provide evidence of the different types of variance that we discussed in Section~\ref{introduction}. The analysis here examines how variations in the image data lead to variance in the results of trained ResNet-50 models. In Section~\ref{exp_dataset}, we investigate variance that occurs in the results of classifiers trained and evaluated on different datasets. 
In Section~\ref{long_tail}, we demonstrate how the variance in long-tailed distributions shows similar trends to variance across datasets. In Section~\ref{exp_few_shot}, we conduct few-shot experiments to investigate trends unique to higher-variance scenarios. 
These experiments provide many examples of both intuitive and counter-intuitive trends and types of variance that occur in FGVC results. Together, they make a compelling case for the need to consider variance metrics in addition to accuracy metrics, a need we further emphasize in Section~\ref{var_by_method}.

\begin{table}[ht]
\begin{center}
\begin{tabular}{|l|c|c|c|} 
 \hline
 Dataset & Classes & \# Images & Accuracy  \\
 \hline
Aircraft & 100 & 10000 & 84.61 \\
Birds & 200 & 11788 & 81.46  \\
Cars & 196 & 16185 & 89.86  \\
Dogs & 120 & 28580 & 82.22  \\
NABirds & 555 & 48000 & 78.58 \\
 \hline
\end{tabular}
\end{center}
\caption{Baseline performance (averaged over 10 trials) of our ResNet-50s on different FGVC datasets. This is not intended to be competitive with state-of-the-art results, but instead gives benchmarks for our models. From this table onward, ``Birds'' refers to CUB.}
\label{table:baselines}
\end{table}

\phantom{foo}

\textbf{Training Procedure}. Unless otherwise specified, we use the same network architecture for each of these experiments -- a ResNet-50 pre-trained on ImageNet~\cite{DBLP:journals/corr/ZeilerF13}. We use the implementation available in PyTorch~\cite{paszke2017automatic}. We optimize cross-entropy loss using the Adam optimizer~\cite{2014arXiv1412.6980K}, starting with a learning rate of 0.0001. We train for 50 epochs, decreasing the learning rate by a factor of 10 at 15, 25, and 35 epochs. During training, we apply random horizontal flips and take random re-sized crops (224 x 224). During evaluation, our networks take center crops. To place our method in context, the accuracy scores of this architecture for different popular fine-grained datasets (FGVC datasets) are provided in Table~\ref{table:baselines}.

\textbf{Terminology}. We use the term \textit{per-class accuracy} to refer to the classification accuracy of the subset of images belonging to one particular class rather than the whole dataset. For CUB, which has 200 classes, this means that there are 200 corresponding per-class accuracies: 1 for each class. \textit{Overall deviation} refers to the standard deviation of all per-class accuracy scores for a particular experiment. For example, if 10 models are trained and evaluated on CUB, which has 200 classes, then the \textit{overall deviation} is the standard deviation of the 2000 resulting per-class accuracies  \textit{Per-class deviation} is the standard deviation of per-class accuracy across multiple training runs, averaged over all classes. For the experiments here, where each data point is the result of 10 training runs, the \textit{per-class deviation} is computed by taking the standard deviation of the 10 accuracy scores for a given class, for all classes, and averaging those standard deviations. Whereas overall standard deviation describes the range of accuracy values and their relationship to the mean, per-class standard deviation describes the differences in performance between the models for each individual class.

\subsection{Variance by Visual Domain}\label{exp_dataset}

\begin{figure}[ht]
    \centering
    \includegraphics[width=\linewidth]{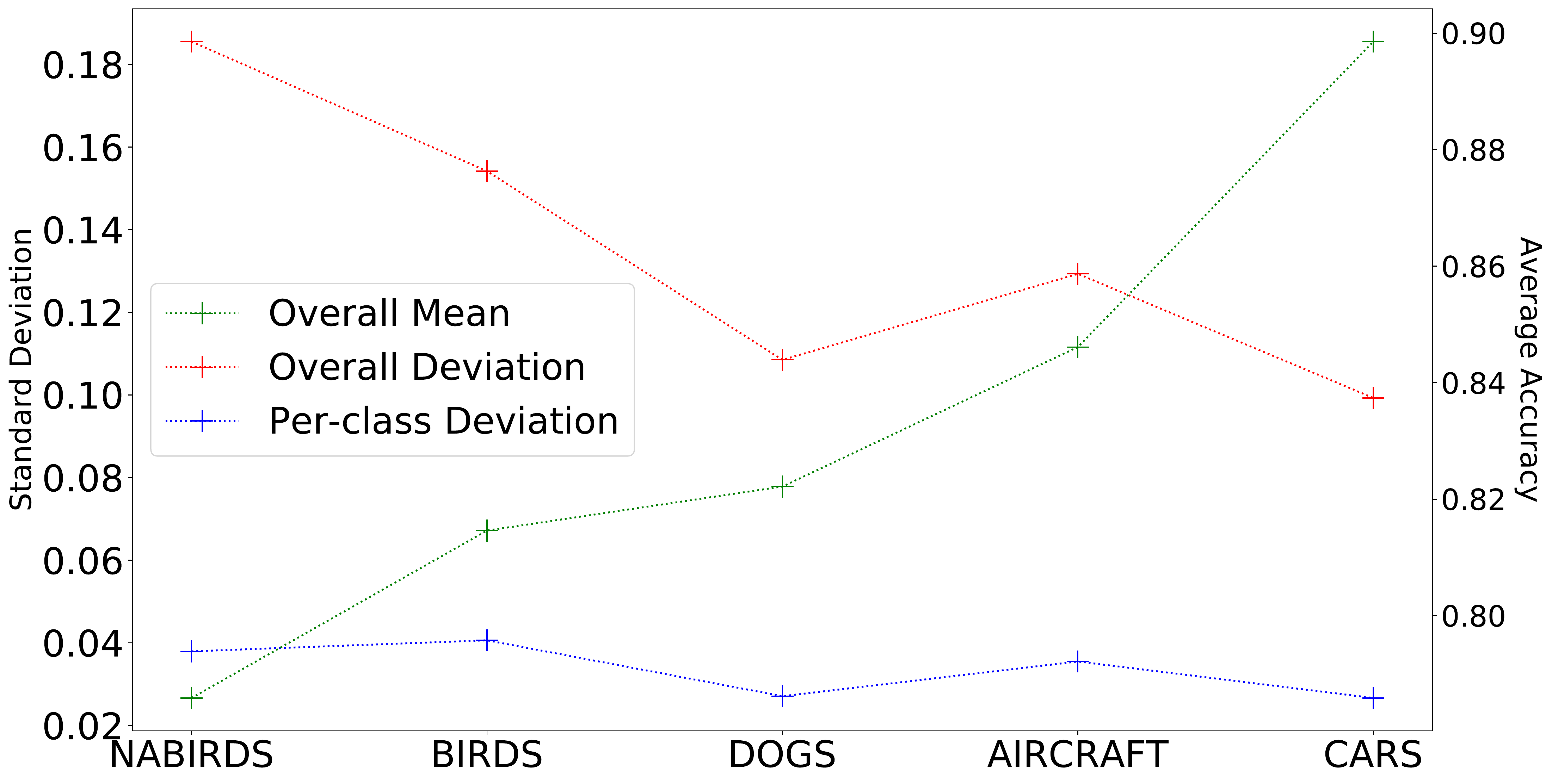}
    \caption{Overall standard deviations, average per-class standard deviations, and overall average accuracies for different datasets. Datasets are sorted by overall accuracy.}
    \label{dataset_stds}
\end{figure}

We train and test 10 networks on each of the FGVC Aircraft~\cite{MajiRKBV_arXiv2013}, Caltech-UCSD Birds, Stanford Cars~\cite{KrauseSDF_ICCVW2013}, Stanford Dogs~\cite{KhoslaJYF_CVPRW2011}, and North American Birds~\cite{VanHornBFHBIPB_CVPR2015} datasets, which we refer to as Aircraft, Birds, Cars, Dogs, and NABirds, respectively. More information on the datasets can be found in Table~\ref{table:baselines}. The results are summarized in Figure~\ref{dataset_accs} and Figure~\ref{dataset_stds}.

Clear patterns emerge in Figure~\ref{dataset_accs}. For each dataset, the model tends to do fairly well, correctly labeling more than 75\% of all test images. Notice that the distribution of per-class accuracies vary from 50\% to 100\%, with most accuracies closer to the maximum than the minimum. Critically, most per-class accuracies are above the mean. While it may seem beneficial that the average class is labeled more accurately than the overall accuracy conveys, this is actually a major issue. What this means, and Figure~\ref{dataset_accs} represents this as well, is that the model performs quite poorly for a non-trivial subset of classes. Taking Birds as an example, for 25 species, the model fails to accurately label more than 60\% of the images. 

This demonstrates the importance of using metrics besides accuracy. While top-1 accuracy conveys how the model performs on average, it fails to communicate the majority of the information shown in Figure~\ref{dataset_accs}. Overall accuracy communicates little about the minimum and maximum accuracy, and nothing about the distribution of accuracies between those two values. Overall deviation, on the other hand, expresses much of that information with a single number. While it does not communicate in concrete terms the maximum or minimum, when combined with overall accuracy it can be used to infer much about the shape of the per-class accuracy curve, including how far from the mean individual per-class accuracies would be. Indeed, as expected from the graphs in Figure~\ref{dataset_accs}, overall deviations are fairly high for each dataset, ranging from just under 10\% for Cars to nearly 20\% for NABirds.

Another finding, from Figure~\ref{dataset_stds}, is that there seems to be a correlation between overall standard deviation and per-class standard deviation, meaning that when the per-class accuracies are distributed closer to the mean, they also tend to vary less between trained instances of a given model. Additionally, accuracy seems to have an inverse correlation with standard deviation, where the datasets with better overall accuracy also have 1) tighter distributions and 2) less disagreement between models.

Some aspects of the variance seem to contradict intuition. For example, as Figure~\ref{dataset_accs} shows, the accuracy for any given class can vary significantly across different training runs, despite the fact that \textit{they all start from the same pre-trained weights}. With NABirds, for instance, there were classes for which one ResNet-50 would label images 35\% more accurately than another ResNet-50. Another surprising finding is that while it has neither the highest nor lowest number of classes, and also has neither the highest nor lowest number of images per class, the ResNet-50s we trained achieved their highest accuracies and lowest deviations (both overall and per-class) on Cars, which suggests the dataset is inherently easier than some of the others. Dogs, for example, has more images representing each class and approximately 60\% as many categories, but the yet our ResNet-50s achieve both less consistent and less accurate results on Dogs than on Cars. 

\subsection{Variance with Long-tailed Distribution}\label{long_tail}

\begin{figure}[t]
    \centering
    \includegraphics[width=\linewidth]{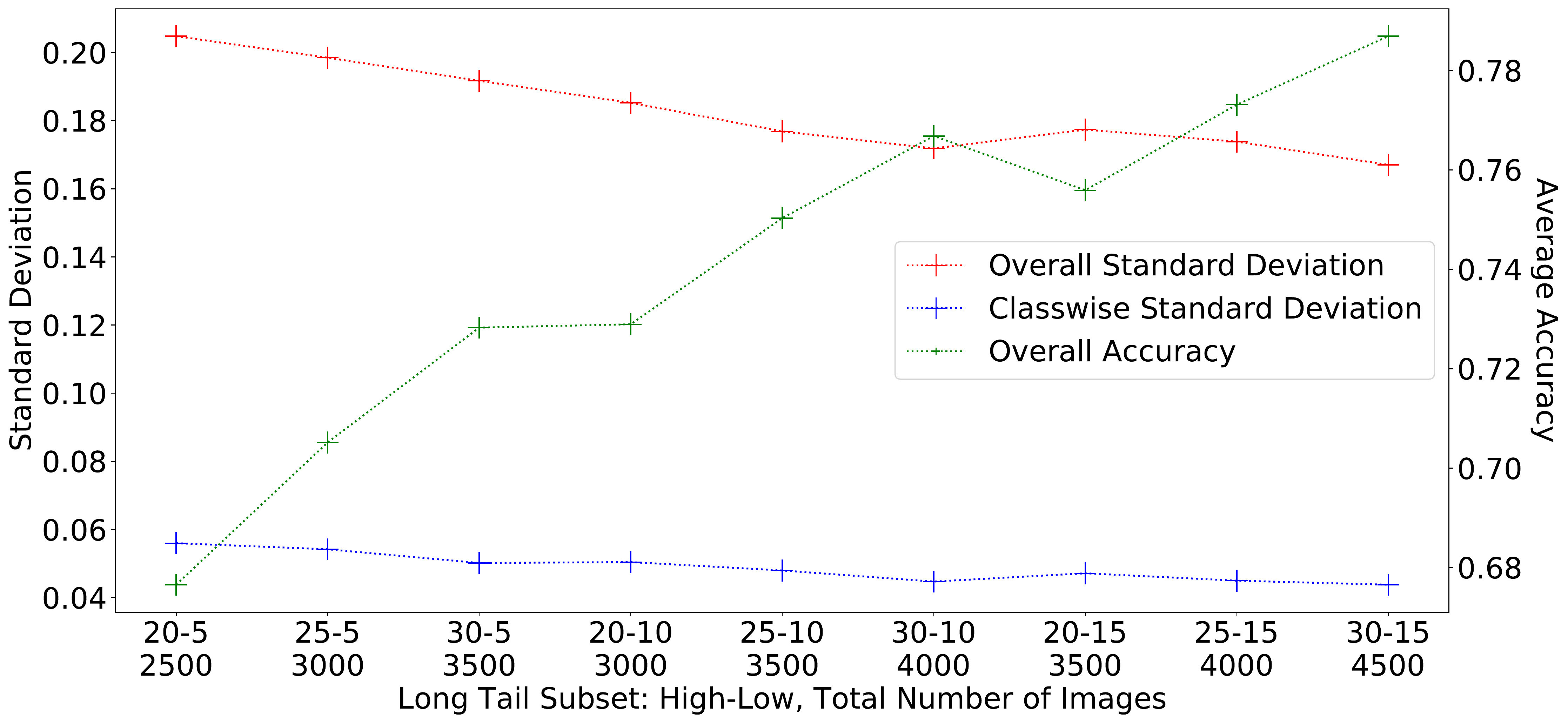}
    \caption{Standard deviations and accuracy results for different long-tailed subsets of the Birds dataset. The subsets take a linearly decreasing number of images from each class, from the high number (left of dash) to the low number (right of dash). Total number of images is shown beneath the dash.}
    \label{long_tail_info}
\end{figure}

\begin{figure*}[t!]
\centering
\includegraphics[width=\linewidth]{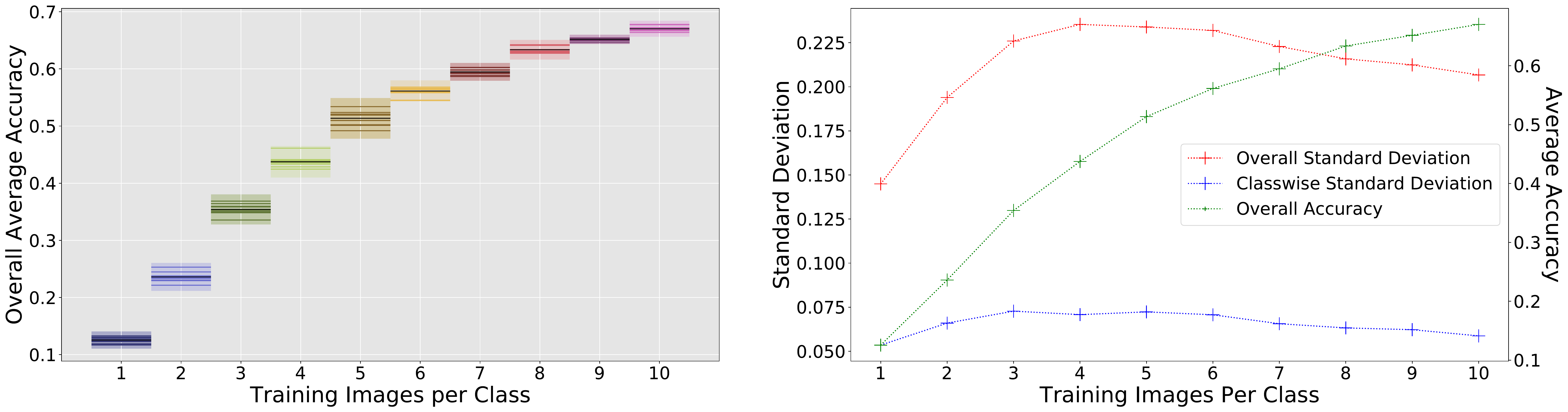}
\caption{\textbf{Left}. Follows the convention described in Figure~\ref{optim_ticks_combined}. Accuracy for few-shot subsets of the Birds dataset.  \textbf{Right}. Standard deviation and accuracy for the same few-shot subsets.}
\label{few_shot_results}
\end{figure*}

Long-tailed distributions were simulated by taking subsets of the Birds dataset, where for a given distribution, the subset was initialized randomly and then kept the same for all trials. These subsets involved a ``high'' number, which represents the maximum number of images per class for the subset, and a ``low'' number, which represents the minimum number of images per class. Classes were then randomly assigned numbers of images so that all numbers from low to high number were used evenly. Additionally, numbers were adjusted when possible so that the total number of images would be multiples of 100. Thus, the 20-5 (high number is 20, low number is 5) distribution had the lowest number of total images, at 2500, and 30-15 had the most, at 4500.

From Figure~\ref{long_tail_info}, note that while accuracy tends to improve as the number of images increases, that result is strongly dependent on the distribution of images, to the point where the 20-15 distribution is almost 4\% better than the 30-5 distribution in terms of average accuracy, even though they both have 3500 images. 
The deviations follow the same trends, but in reverse- they tend to decrease as accuracy increases.
Figure~\ref{long_tail_info} thus reinforces a key trend from Figure~\ref{dataset_stds} -- the correlation between accuracy, overall deviation, and per-class deviation. There is only one minor deviation from this tendency in the whole chart, when per-class deviation increases slightly from the 30-5 subset to the 20-10 subset while accuracy increases and overall standard deviation decreases.

Interestingly, both the per-class and overall deviations in Figure~\ref{long_tail_info} decrease consistently not as the overall number of images increases, but more so as the minimum number increases. This is likely due to the fact that underrepresented classes tend to be classified both less accurately (affecting overall deviation) and less consistently between training runs (affecting per-class deviation).

\subsection{Variance In Few-Shot}\label{exp_few_shot}

10 subsets of Birds were taken, with the number of training images limited from 1 up to 10 images per class, inclusive. We trained 10 models on each subset and averaged the results. From the graph on the left in Figure~\ref{few_shot_results}, we can see that the variance in overall results changes dramatically from 1 to 10 images per class, with the distributions getting much tighter as the number of training images increases. Note that for these few shot trials, overall accuracy results can vary by multiple percentage points between trials. 

The graph on the right of Figure~\ref{long_tail_info} is even more interesting, as it contradicts some of the trends observed in Section~\ref{exp_dataset} and in Section~\ref{long_tail}. We see that overall and per-class deviations are both very high, as could be expected in a lower accuracy setting. However, variance does not decrease monotonically as accuracy increases. In fact, The variance actually increases with accuracy until the jump from 4 to 5 images per class. This calls into question whether variance is as straight-forward as much of this section has suggested it may be (varying inversely with accuracy), or if it can change independent of accuracy, and thus can be manipulated directly, similarly to how researchers currently strive to manipulate (increase) accuracy. Section~\ref{var_by_method} answers this question, by making it clear that variance exhibits a significant degree of independence from accuracy.

%
%
\section{Variance by Method} \label{var_by_method}
Whereas Section~\ref{var_by_data} focuses on variance resulting from different attributes of the dataset, this section is dedicated to assessing the impact of variance in various methods for fine-grained visual categorization. In Section~\ref{exp_optimizer}, we examine popular baseline optimizers and model architectures to highlight the importance of considering variance in both designing and reporting the results of FGVC pipelines. In Section~\ref{exp_loss} we analyze the effect different loss functions have on variance. Finally, in Section~\ref{exp_sota} we present our findings on variance in current state-of-the-art methods. From our discussions on loss functions and state-of-the-art algorithms we identify techniques that seem to affect variance independent of accuracy, and identify necessary future work.

\subsection{Variance by Optimizer and Architecture}\label{exp_optimizer}

\begin{figure}[t!]
\centering
\includegraphics[width=\linewidth]{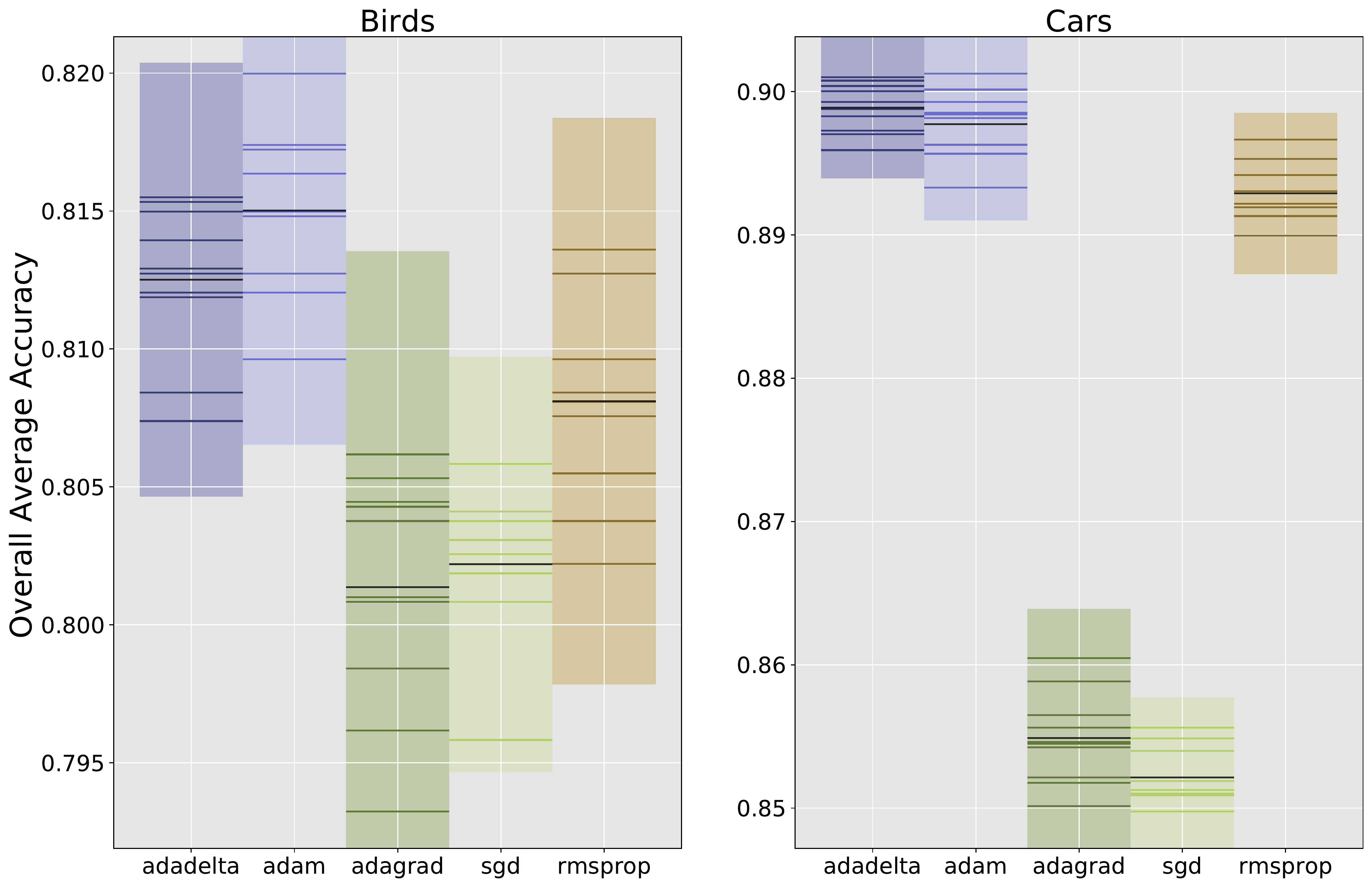}
\caption{Overall accuracy for the Birds and Cars datasets trained with different optimizers. Colored lines show the accuracy results from each of the 10 trials. Black lines show the mean across the 10 trials. The shaded region represents 3 standard deviations above and below this mean.}
\label{optim_ticks_combined}
\end{figure}

For optimizers, we compare Adagrad~\cite{Duchi:2011:ASM:1953048.2021068}, Adadelta~\cite{DBLP:journals/corr/abs-1212-5701}, Adam (used in the other experiments), RMSprop~\cite{DBLP:journals/corr/Graves13}, and stochastic gradient descent (SGD).
For models, we compare DenseNet-161~\cite{DBLP:journals/corr/HuangLW16a}, ResNet-152~\cite{DBLP:journals/corr/HeZRS15}, VGG19~\cite{SimonyanZ_ICLR2015} with batch normalization, and Inception v3~\cite{DBLP:journals/corr/SzegedyVISW15}.  

Figure \ref{optim_ticks_combined} shows the results of the experiments with optimizers. The performance of Adadelta and Adam is similar. Nevertheless, it is worth noting that Adadelta has a lower standard deviation and a higher mean for both Birds and Cars, while Adam has the higher max for Cars. Also worth noting is that Adam's \textit{maximum} accuracy is higher than Adadelta's \textit{average} accuracy for both datasets. This could result in the reporting problem described in Section \ref{introduction} -- a researcher using Adam could make it appear as though Adam has the ``better'' result by reporting only the maximum (on Birds) or the average of the best 5 trials (on Cars) instead of all 10.

\begin{figure}[t!]
\centering
\includegraphics[width=\linewidth]{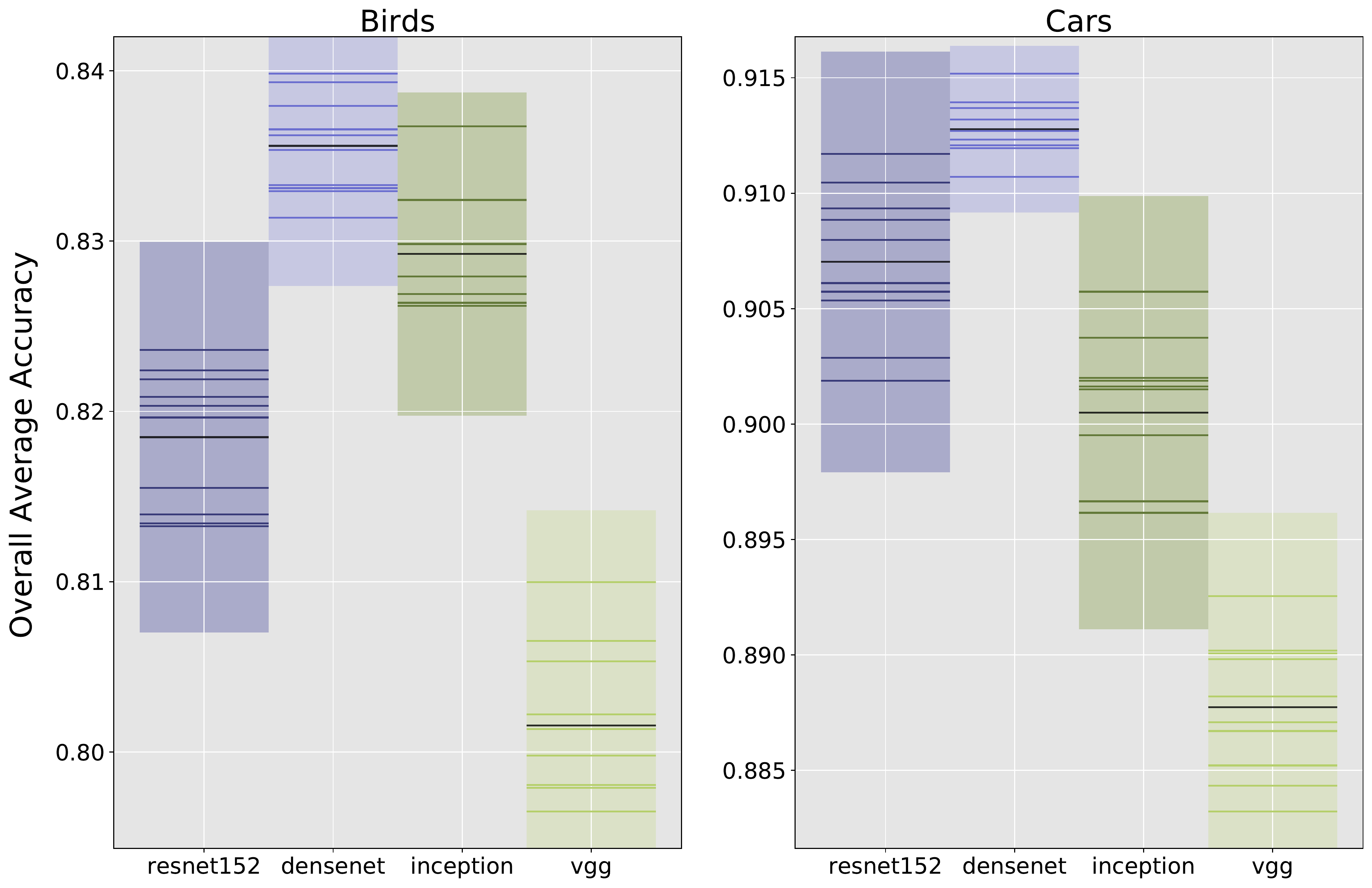}
\caption{Overall model accuracy for different models trained and tested on the Birds and Cars datasets. Uses format explained in caption for Figure \ref{optim_ticks_combined}.}
\label{model_ticks_combined}
\end{figure}

Figure \ref{model_ticks_combined} shows the variance in overall accuracy for different models trained and evaluated on Birds and Cars. This offers more evidence to support the importance of reporting an average over more than 5 trials, without cherry-picking. With DenseNet-161, for example, while its average accuracy for Birds is higher than the average of the 10 Inception-v3 networks, it is lower than the overall accuracy from 1 of the Inception-v3 networks. In a similar fashion, the 5 lowest accuracies for ResNet-152 on Cars overlap with the highest 5 for Inception-v3. 

Surprisingly, the networks with the highest accuracy scores do not always have the most consistent scores. While the best network by overall accuracy, the DenseNet-161, has the lowest standard deviation, the worst network, the VGG-19, seems to be on par with ResNet-152 and Inception-v3 in terms of consistency, despite having a significantly worse overall accuracy. This highlights the need for comparing results across multiple runs, and is an additional consideration to take into account when selecting a network type, especially when resources are limited and ensemble methods are not feasible.

\subsection{Variance by Loss Function}\label{exp_loss}

\begin{figure}[t!]
\centering
\includegraphics[width=\linewidth]{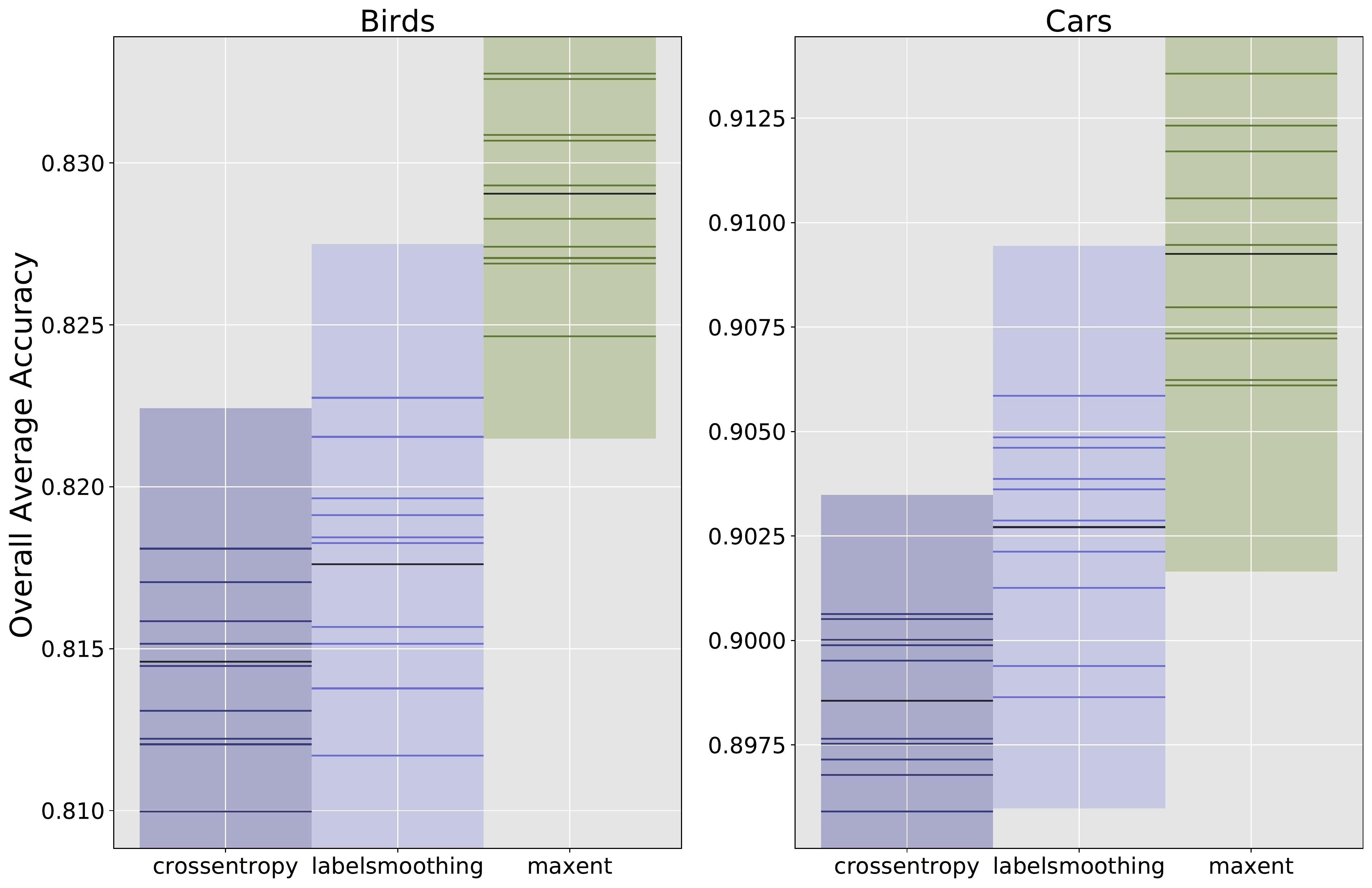}
\caption{Overall model accuracy for models trained with different loss functions and tested on the Birds and Cars datasets. Uses format explained in caption for Figure~\ref{optim_ticks_combined}.}
\label{loss_ticks_combined}
\end{figure}

\begin{table}[ht]
\begin{center}
\begin{tabular}{|lccc|}
\hline
 & & \multicolumn{2}{c|}{Deviation} \\
Loss Name & Accuracy & Overall & Per-class \\
\hline
\multicolumn{4}{|l|}{\textbf{CUB}} \\
baseline       & 81.46\% & 0.1539 & 0.03962 \\
labelsmoothing & 81.79\% & 0.1555 & 0.03942 \\
maxentropy         & 82.90\% & 0.1493 & 0.03556 \\
\hline
\multicolumn{4}{|l|}{\textbf{Cars}} \\
baseline       & 89.89\% & 0.0987 & 0.02632 \\
labelsmoothing & 90.26\% & 0.1007 & 0.02500 \\
maxentropy         & 90.96\% & 0.0978 & 0.02418 \\
\hline
\end{tabular}
\end{center}
\caption{Presents accuracy and variance results for 10 ResNet-50s each trained with 3 different loss functions, for CUB and Cars.}
\label{table:loss_results}
\end{table}

Recent work has challenged the notion that one-hot targets with cross-entropy are the best way to model the objective for FGVC. We specifically compare 2 different approaches to standard one-hot cross-entropy: label-smoothing~\cite{muller2019does,SzegedyVISW_CVPR2016}, which treats the target class as most probable while assigning a uniform, non-zero probability to all other classes, and maximum entropy regularization~\cite{DubeyGRN_NeurIPS2018}, where regular cross entropy is computed and the entropy of each prediction vector is subtracted from it to calculate the overall loss.

The results are somewhat unexpected. While Figure~\ref{loss_ticks_combined} shows that maxentropy outpeforms labelsmoothing which outperforms baseline crossentropy, as expected, note that the accuracies for label smoothing are distributed over a wider range than those of the baseline. In a related vein, Table~\ref{table:loss_results} shows that overall deviation (which describes the distribution of per-class accuracies around the mean) is larger for labelsmoothing than the baseline on both datasets, which is quite different from the trend we observed in Section~\ref{var_by_data}. The existence of a method with higher accuracy and also higher variance provides evidence for the idea that variance can be targeted and solved as its own goal, independent of accuracy. Future work in loss functions could potentially focus on lowering the variance.

\subsection{Variance in the State-of-the-Art}\label{exp_sota}

\begin{figure}[t!]
\centering
\includegraphics[width=\linewidth]{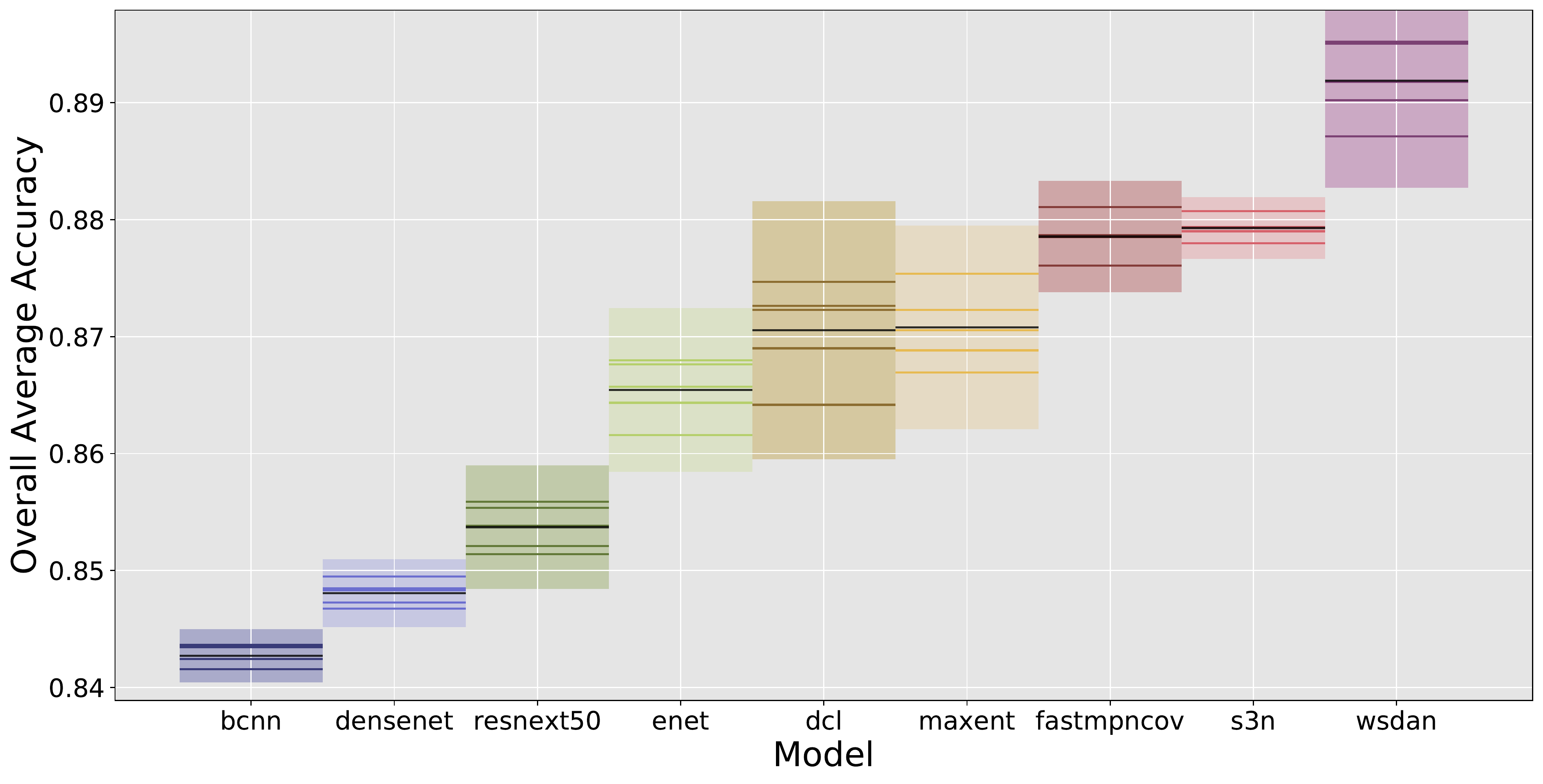}
\caption{Overall model accuracy for different models trained and tested on CUB. Uses format explained in caption for Figure~\ref{optim_ticks_combined}}
\label{sota_overall_ticks}
\end{figure}

\begin{figure}[t!]
\centering
\includegraphics[width=\linewidth]{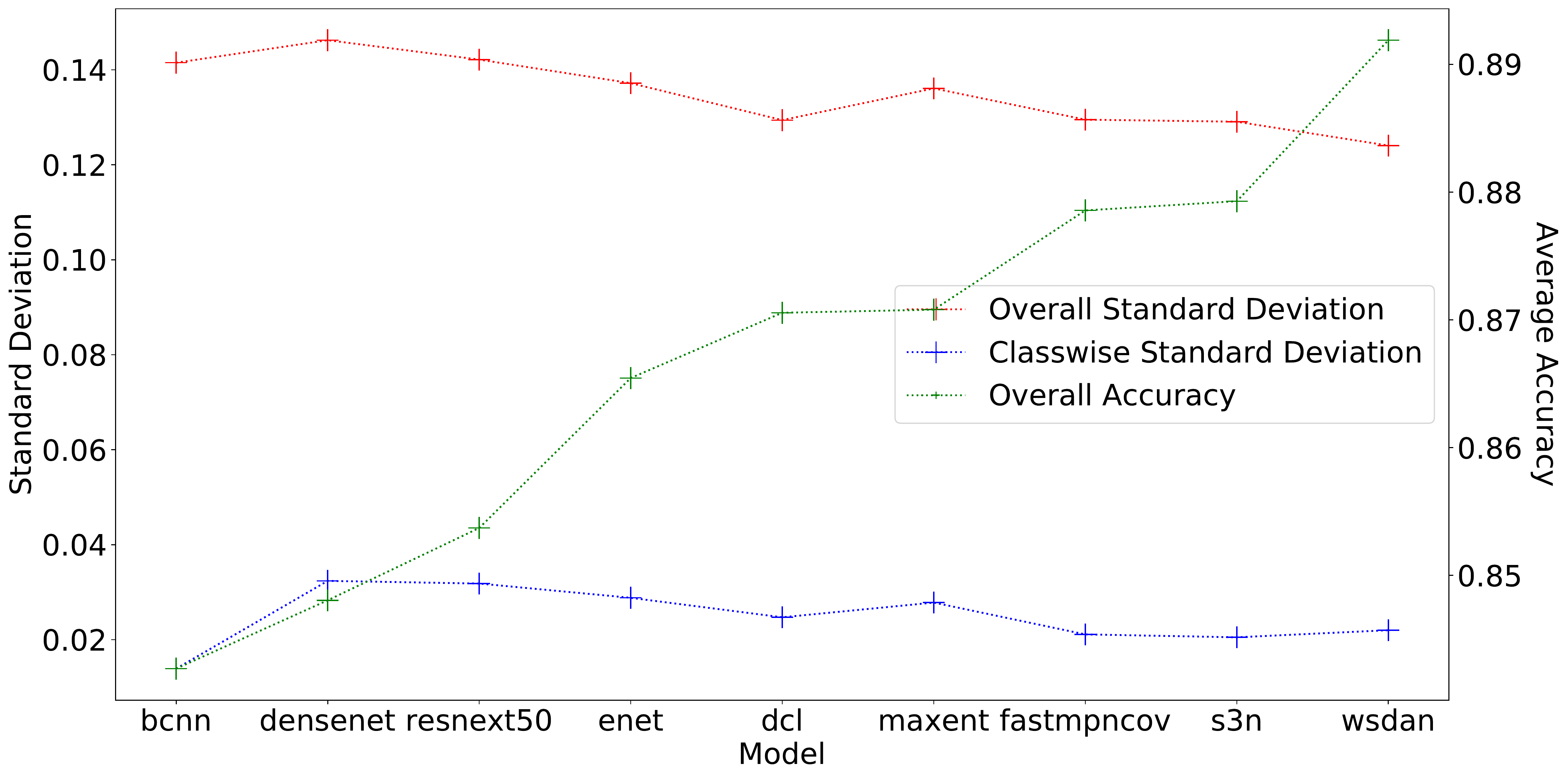}
\caption{Accuracy and deviations for the SOTA models.}
\label{sota_stds}
\end{figure}

For state-of-the-art methods, we consider several different high-performing approaches: WS-DAN~\cite{HuQHL_arXiv2019}, Fast MPN-COV~\cite{li2017second}, DCL~\cite{ChenBZM_CVPR2019}, S3N~\cite{DingZZYJ_ICCV2019}, Densenet161~\cite{HuangLW_CVPR2017_DenseNet}, Densenet161 with maximum entropy regularization, EfficientNet~\cite{tan2019efficientnet}, BilinearCNN, and ResNeXt50~\cite{xie2017aggregated}. Unlike in previous sections, we only trained 5 of each of these, and we evaluate on CUB only. Additionally, we used larger images -- 448x448.

One of the most surprising aspects of Figure~\ref{sota_overall_ticks} is what it shows about the deviation in overall accuracy, particularly in how it does not decrease as accuracy increases. High performing methods such as DCL and WSDAN show large variations in overall accuracies between the 5 trials. On the other hand, the lowest performing method, BCNN, and the second highest, S3N, both have very tight overall accuracy distributions. It is clear that there is something about the design of the models that lends itself to more consistent performance, and insofar as it is not always practical to use an ensemble, there is a real possibility one could get an unlucky draw with models such as DCL and WSDAN. It is therefore important for researchers to consider the consistency of their models, and important for the community to be aware of how methods stack up in that regard.

The deviations in Figure~\ref{sota_stds} also do not follow the patterns from the previous sections. The one trend that is mostly preserved is the correlation between the two deviations, though even that is violated with WSDAN, where overall deviation decreases while per-class deviation increases. One of the more interesting violations of the accuracy-deviation correlation is maximum entropy, which is comparable in variance to enet, which it outperforms by more than 0.5\%. The transition from BCNN to densenet is similar; while accuracy increases considerably, the deviations also both increase substantially. 

Both figures indicate that Bilinear CNN is very unique in terms of its variance-accuracy relationship. In particular, despite having the lowest accuracy, it has a low per-class deviation. Additionally, there is little variance in the overall accuracy results of BCNN across its 10 trials. Unsurprisingly, it does not have the lowest overall deviation -- methods whose accuracies are closer to 100\% will naturally have lower overall deviations. These results prove that variance is not completely dependent on accuracy, and some methods, such as BCNN, are more consistent than methods that are otherwise considered ``better,'' such as DCL. Studying these differences could be the key to minimizing variance in the highest performing SOTA methods.

%
%
\section{Conclusion} \label{conclusion}
As incremental improvements to FGVC techniques increase the applicability of automated image classification to real world problems, it becomes increasingly important to cultivate a deep understanding of how these methods perform, both in absolute and relative terms. The types of variance demonstrated in this paper reveal an unexplored side of the FGVC problem. Specifically, this work investigates variance by demonstrating its prevalence and highlighting the need for change. We give examples of how to use deviation to quantify different types of variance using standard deviation calculations. We demonstrate when intuitive assumptions about the relationship between accuracy and variance apply, and when they fail. We promote the adoption of variance as a problem, and we argue that minimizing variance is a goal worthy of consideration as researchers continue to develop increasingly accurate FGVC methods.

\vspace{.1in}

\section*{Acknowledgments}
\noindent
This work was supported by the National Science Foundation under Grant No. IIS1651832. 
We gratefully acknowledge the support of NVIDIA Corporation for their donation of multiple GPUs that were used in this research.

\vspace{.25in}

%
%

\balance

\bibliography{main.bbl}
\bibliographystyle{ieee}
\end{document}